\documentclass{article} 
\usepackage[final]{colm2026_conference}

\usepackage{microtype}
\usepackage{hyperref}
\usepackage{url}
\usepackage{booktabs}
\usepackage{wrapfig}
\usepackage{amsmath}
\usepackage{amssymb}
\usepackage{mathtools}
\usepackage{amsthm}
\usepackage{makecell}
\usepackage{graphicx}
\usepackage{subcaption}
\usepackage{xcolor}
\usepackage{multirow}
\usepackage{multicol}
\usepackage{pifont}
\usepackage[misc]{ifsym}
\usepackage[most]{tcolorbox}

\usepackage{color}
\usepackage[table]{xcolor}
\definecolor{mygreen}{RGB}{0,120,90}

\usepackage[capitalize,noabbrev]{cleveref}

\definecolor{darkblue}{rgb}{0, 0, 0.5}
\hypersetup{colorlinks=true, citecolor=mid, linkcolor=primary, urlcolor=mid}

\newtcolorbox{mybox}[2][]{
  colbacktitle=red!10!white, 
  colback=blue!10!white,
  coltitle=red!70!black, 
  title={#2},
  fonttitle=\bfseries,
  #1
}
\newtcblisting{promptbox}[2][]{%
  colbacktitle=red!10!white, 
  colback=blue!10!white,
  coltitle=red!70!black, 
  title={#2},
  fonttitle=\bfseries,
  width=\textwidth,   
  listing only,                           
  listing engine=listings,                
  listing options={
    basicstyle=\ttfamily\small,          
    breaklines=true,                      
    breakindent=0pt,                      
    columns=flexible
  },
  #1
}

\newcommand{\E}{\mathbb{E}}
\newcommand{\KL}{D_{\mathrm{KL}}}

\newcommand{\train}{\mathcal{D}}

\renewcommand{\cite}[1]{\citep{#1}}

\title{MemTrain: Self-Supervised Context Memory Training}

\author{
\mbox{
Ziheng Li$^{1,2\dagger}$, Xingrun Xing$^{2\dagger}$, Haoqing Wang$^{2}$,
}\\
Zhi-Hong Deng$^{1{~\textrm{\Letter}}}$, and Yehui Tang$^{2{~\textrm{\Letter}}}$ \\
$^1$ State Key Laboratory of General Artificial Intelligence, School of Intelligence Science and Technology, Peking University\\
$^2$ Samsung Research, Beijing, China\\
\texttt{\{liziheng,zhdeng\}}@pku.edu.cn\quad\texttt{yehui.tang}@samsung.com \\
$^\dagger$~Equal Contribution\quad\quad$^\textrm{\Letter}$~Corresponding Author
}

\begin{document}

\maketitle
\begin{abstract}
Memory is an indispensable capability for long-horizon LLM agents, enabling them to preserve and utilize information accumulated across extended interactions.
Existing memory-agent approaches are typically trained end-to-end with reinforcement learning on downstream tasks.
However, collecting high-quality annotated problems for memory-intensive scenarios is costly, and the resulting training data often lack sufficient diversity to cover general memory behaviors. 
In this work, we propose MemTrain, a self-supervised training framework for generally enhancing the context-memory capability of LLM agents for more effective downstream post-training.
MemTrain introduces two coupled proxy tasks over unlabeled Wikipedia corpora:
(1) an end-to-end masked reconstruction objective, which requires the model to recover masked entities after multiple rounds of memory updates, thereby encouraging memory maintenance from the final outcome perspective; and
(2) an intermediate memory recall objective, which requires the model to reconstruct masked historical information using intermediate memory states, encouraging faithful compression and memory completeness throughout the interaction process.
The two objectives are jointly optimized using GRPO. 
Extensive experiments on long-text QA and search-based QA benchmarks demonstrate that MemTrain consistently improves downstream memory-intensive reasoning performance across different models, achieving gains of up to 17.67 points over direct task-specific post-training.
\end{abstract}

\section{Introduction}

Large language models (LLMs) have rapidly evolved into increasingly capable agents that can reason, plan, and interact with external environments~\citep{singhOpenAIGPT5System2025,teamKimiK2Open2025, deepseek-aiDeepSeekR1IncentivizingReasoning2025}.
However, a key bottleneck for long-horizon agentic tasks is \emph{memory}: the ability to preserve and utilize information acquired many turns earlier.
In realistic interactive settings, an agent continuously receives new observations, generates intermediate thoughts, and must maintain relevant past information across turns.
A straightforward solution is to append the full interaction history into the prompt~\citep{yaoReActSynergizingReasoning2023a}, but this quickly becomes prohibitively expensive as the trajectory grows.
Consequently, enabling agents to operate with a \emph{fixed-size persistent memory} remains an important challenge for scalable long-horizon deployment.

Recent work has explored \emph{context memory} agents~\citep{zhou2025mem1,yu2025memagent,yan2025MemoryR1Enhancing,yuan2026MemSearcherTrainingLLMs}, where each interaction round is conditioned on a compact memory state rather than the entire history.
At turn $t$, the model receives an input of the form $[\texttt{memory}_{t-1}; \texttt{input}_t]$, produces a response, and updates the memory into $\texttt{memory}_t$. This paradigm allows near-constant context usage while preserving historical information, and can be optimized end-to-end within the language model itself.
However, existing memory agents are typically trained using reinforcement learning with verifiable reward (RLVR) on downstream tasks. Such approaches require expensive labeled data, making it difficult to obtain sufficiently diverse training data that covers the wide range of memory behaviors. 
Consequently, memory capabilities learned in this manner are often domain-specific and exhibit limited generalization. These limitations highlight the need for a general-purpose self-supervised training paradigm.

Meanwhile, recent advances in reasoning have explored reinforcement learning with pre-training data\citep{dongReinforcementPreTraining2025,liReinforcementLearningPreTraining2025,xingPretrainZeroReinforcementActive2025}.
They construct self-supervised proxy tasks over unlabeled corpora by chain-of-thought next-token prediction to generally improve the reasoning ability.
However, memory learning poses distinct challenges from reasoning.
The memory target is inherently latent and process-dependent, as the model must continuously decide what information to preserve, compress, and recall over time.
Consequently, designing a proxy task that faithfully captures the underlying memory mechanism remains a significant challenge.

To address this challenge, we propose \textbf{MemTrain}, a self-supervised training framework for improving the general context-memory capability of LLM agents in order to better support downstream post-training.
MemTrain is built upon two coupled proxy tasks constructed from Wikipedia passages:
(1) an end-to-end masked reconstruction task, which requires the model to recover masked entities after multiple rounds of memory updates, thereby encouraging effective memory maintenance and utilization; and
(2) an intermediate memory recall task, which requires the model to reconstruct additional masked entities from earlier interaction history using intermediate memory states, encouraging memory completeness and faithful compression throughout the memory update process. The two objectives are jointly optimized with GRPO.
Extensive experiments show that MemTrain consistently improves downstream long-text QA and search-based QA performance over direct task training. The average improvements reach 5.17 points and 10.58 points respectively on Qwen3-4B-Instruct-2507 and reach 17.67 and 8.50 points on Qwen2.5-7B-Instruct.

Our contributions are summarized as follows:
\begin{itemize}
    \item We propose MemTrain, the first self-supervised training framework designed to generally improve the context-memory capability of LLM agents for effective downstream post-training.
    \item We introduce a novel memory-oriented proxy training paradigm that jointly provides outcome-level and process-level supervision signals for memory generation and utilization.
    \item Extensive experiments on long-text QA and search-based QA tasks demonstrate that MemTrain consistently improves downstream post-training performance ceiling on both 4B and 7B models.
\end{itemize}

\section{Related Works}

\paragraph{Memory for Long-Horizon LLM Agents.}
The most widely adopted memory management strategy for LLM agents is to continually append environmental observations and model responses to the context window~\citep{yaoReActSynergizingReasoning2023a}, which is fundamentally limited by the finite context window of LLMs.
To enable unbounded memory, external memory systems have been proposed, where interaction records are compressed or summarized and stored externally.~\citep{yoon2024CompActCompressingRetrieved,li2023CompressingContextEnhance,chhikara2025Mem0BuildingProductionReady,xu2025AMemAgenticMemory}.
\citet{qian2026MemoBrainExecutiveMemory,xu2025AMemAgenticMemory,chen2026RetrieveThinkAgentic} further introduce multi-agent frameworks to support more sophisticated and efficient memory management.
However, external memory systems often overlook the intrinsic synergy between memory and reasoning, while simultaneously increasing overall system complexity.
More recent studies~\citep{zhou2025mem1,yu2025memagent,wu2026ReSumUnlockingLongHorizon,ye2025AgentFoldLongHorizonWeb,yuan2026MemSearcherTrainingLLMs} integrate memory construction and utilization directly into the reasoning process of the agent itself, enabling end-to-end optimization.
Despite their effectiveness, these approaches typically rely on costly task-specific annotations, severely limiting the data diversity.
In this work, we instead propose a self-supervised training framework that enables training on common Internet corpora, significantly enhancing data diversity.

\begin{figure*}[t]
    \centering
    \includegraphics[width=0.95\linewidth]{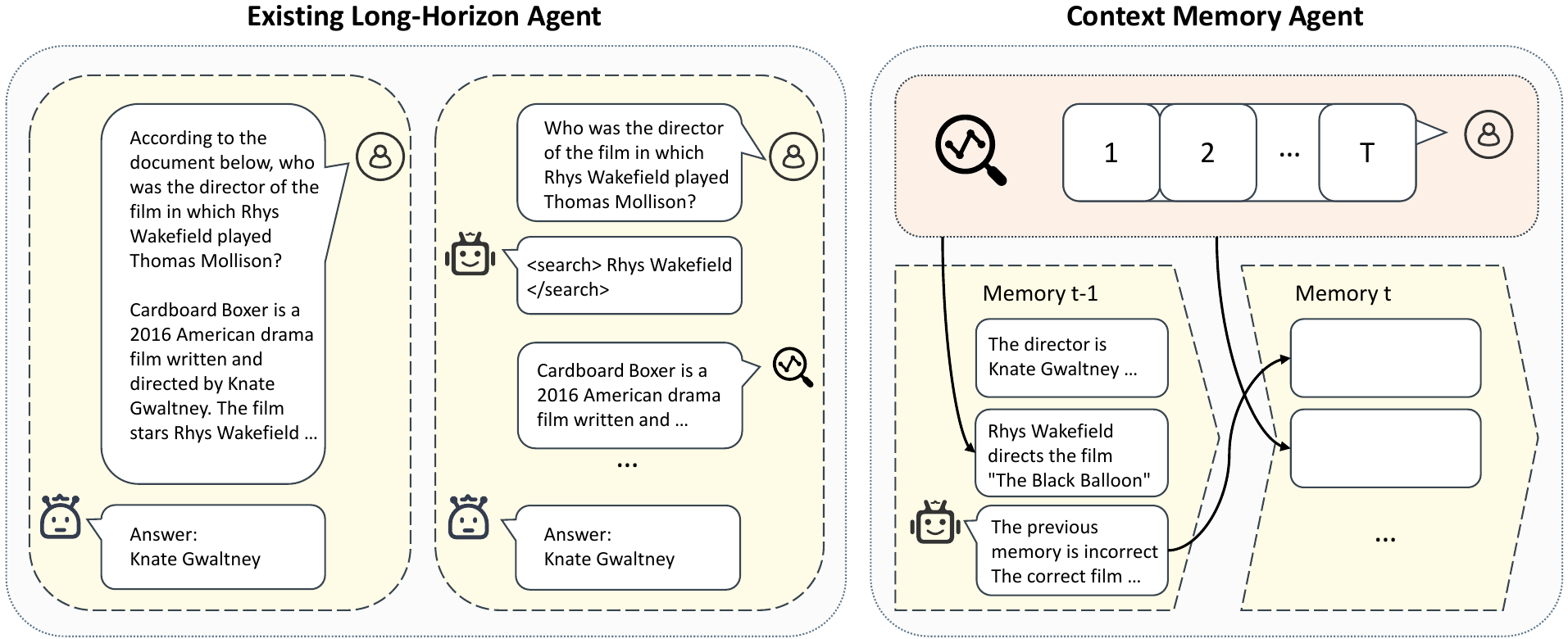}
    \caption{Comparison between existing long-horizon agent and context memory agent. Conventionally, to handle long-context document or multi-turn environment interaction, LLM has to preserve all input in the context, causing high computational cost and attention pressure. By contrast, context memory agent maintains a fixed-length context memory updated at each turn, allowing handle increasing input within feasible resource limit.}
    \label{fig:memory_overview}
\end{figure*}

\paragraph{Reinforcement Learning for LLM Pre-training.}
Reinforcement learning has been extensively adopted during post-training to enhance the reasoning and tool-use capabilities of LLMs~\citep{deepseek-aiDeepSeekR1IncentivizingReasoning2025,yuDAPOOpenSourceLLM2025}. However, post-training methods generally depend on curated question-answer datasets, which limits both scalability and generalization.
Motivated by the success of self-supervised language model pre-training, recent works have explored reinforcement pre-training paradigms that leverage large-scale Internet text.
Quiet-STaR~\citep{zelikman2024QuietSTaRLanguage,huang2025FastQuietSTaR} generates latent rationales at each token position to better predict future text.
RPT~\citep{dongReinforcementPreTraining2025} introduces the next-token reasoning RLVR objective and demonstrates scalable reinforcement learning pre-training for the first time.
RLPT~\citep{liReinforcementLearningPreTraining2025} adopts a similar formulation while incorporating a generative reward model. RLP~\citep{hatamizadeh2025RLPReinforcement} replaces next-token prediction with a contrastive reward to explicitly induce reasoning. 
PretrainZero~\citep{xingPretrainZeroReinforcementActive2025} further proposes an active pre-training framework that synthesizes more informative and valuable training samples. Nevertheless, existing RL-based pre-training approaches primarily focus on single-turn reasoning, leaving the problem of learning effective multi-turn memory maintenance and utilization largely unexplored.

\section{Self-Supervised Memory Training}
In this section, we first formulate the context memory agent (\S~\ref{sec:problem}). We then introduce the two proxy task -- end-to-end masked reconstruction (\S~\ref{sec:e2e}) and intermediate memory recall (\S~\ref{sec:imr}). Finally we describe how we conduct the memory training using GRPO (\S~\ref{sec:optim}).

\subsection{Problem Setup}\label{sec:problem}

Our study is built upon the framework of multi-turn context memory proposed in MemAgent~\citep{yu2025memagent}.
As shown in Figure~\ref{fig:memory_overview}, existing context-memory mechanisms can be abstracted as maintaining a fixed-length memory state $m_t$ at interaction step $t$. At each interaction step, the model receives an input tuple $(m_{t-1}, a_{t-1}, i_t)$, where $a_t$ denotes the action selected by the model at the current step. The action space depends on the target application. For long-context reading agents, actions may correspond to requesting the next text chunk or generating the final answer. For search agents, actions may involve invoking an external search tool or directly returning an answer. For non-terminal actions that interact with the environment, $i_t$ represents the environment input or feedback returned after executing the selected action. Conditioned on $(m_{t-1}, a_{t-1}, i_t)$, the model produces the updated memory state and action, i.e., $(m_{t}, a_{t})$, which are then used in the subsequent interaction step.

Compared with the conventional agent paradigm, where the entire interaction history is continually appended to the context window, context memory maintains a constant context size throughout the trajectory. This design removes the dependence on ever-growing context length, enabling long-horizon interaction beyond the model's native context limit while mitigating attention dilution and avoiding the increasing computational cost associated with long-context processing.

\begin{figure*}[t]
    \centering
    \includegraphics[width=0.95\linewidth]{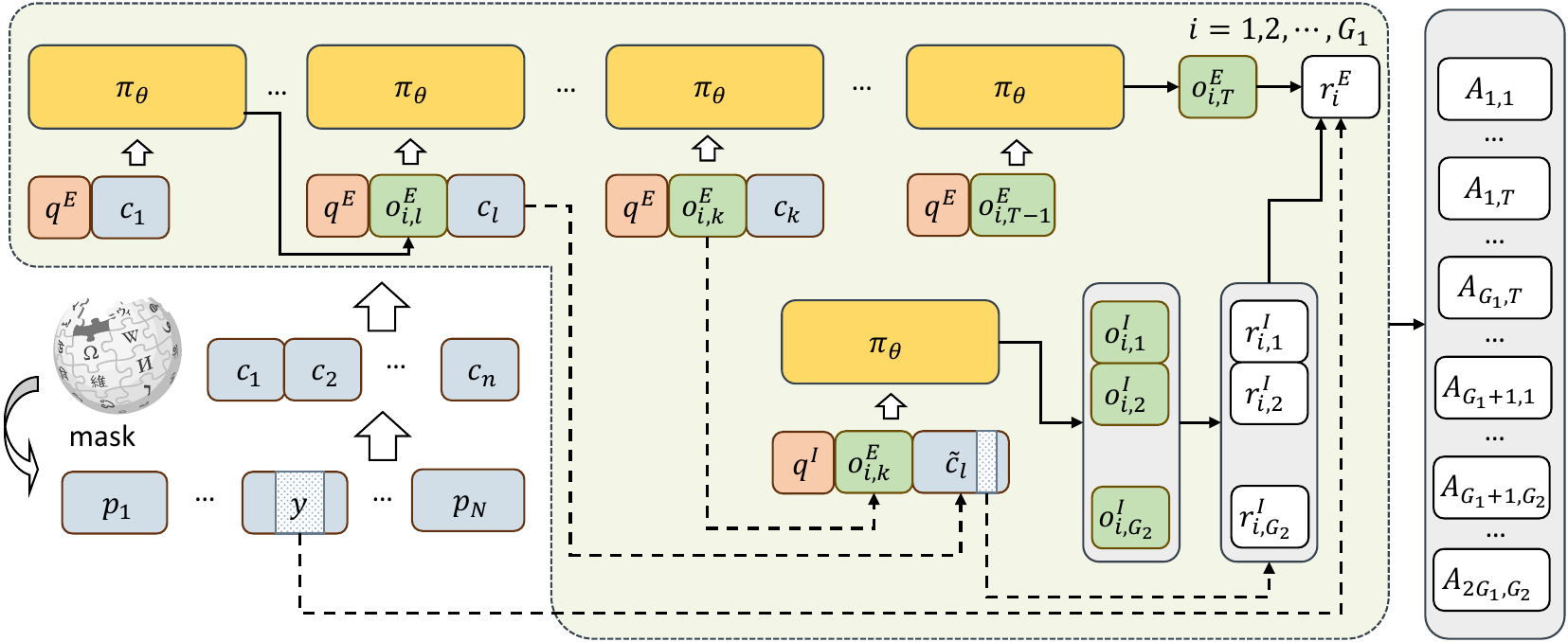}
    \caption{Illustration of MemTrain rollout pipeline during GRPO training. First, we select $N$ passages from the Wikipedia corpus and constructed a chunked input collection $c_{1:T-1}$. Then we sample $G_1$ multi-turn trajectories $o^E_{1:T}$ for recovering masked word $\hat y$ by sequentially reading $c_{1:T-1}$ and update context memory. For each multi-turn trajectory, we randomly select a intermediate memory to recover an input chunk before and generate $G_2$ intermediate memory recall trajectory. Finally, we compute reward and advantage for all $G_1T+G_1G_2$ interactions.}
    \label{fig:grpo}
\end{figure*}

\subsection{End-to-End Masked Reconstruction}\label{sec:e2e}

We construct training samples from raw Wikipedia text. First, we randomly select one passage as the pivot passage. We then retrieve $n_1$ semantically related passages from the corpus together with $N\!-\!n_1\!-\!1$ randomly sampled passages. These $N$ passages are concatenated in random order to form a long document. Next, we randomly select an entity $y$ (e.g., a number or location) from the pivot passage and replace all occurrences of this entity in the document with a special token [MASK].

Following the practice in context-memory research~\citep{yu2025MemAgentReshapingLongContexta}, we segment the long document into fixed-length chunks $\{c_1, c_2, \dots, c_T\}$, where each chunk corresponds to an interaction step. The LLM sequentially processes these chunks to generate a multi-turn trajectory $o_i^E$ (the $i$-th rollout) following
$o_{i,t}^E \sim \pi_\theta(\cdot |q^E, o_{i,t-1}^E, c_t)$,
where $q^E$ denotes the reconstruction prompt detailed in Appendix~\ref{sec:template}.
For $t < T$, the output $o_{i,t}^E$ serves as the context memory for the next interaction step, while $o_{i,T}^E$ denotes the final answer prediction generated solely based on the memory state $o_{i,T-1}^E$, without external input.
Since all occurrences of $y$ are masked, the model cannot simply copy the answer from the document and must instead infer the masked entity through comprehensive long-range information aggregation. This setup provides an end-to-end supervision signal: successful prediction requires preserving and integrating relevant information across multiple memory updates rather than relying on local context alone.

\subsection{Intermediate Memory Recall}\label{sec:imr}

End-to-end rewards alone are often coarse and may not sufficiently constrain the quality of intermediate memory states.
The model may incidentally preserve the information necessary for the final prediction while discarding other important details.
Furthermore, due to error accumulation across multiple interaction steps, optimization based solely on end-to-end outcomes may provide weak and unstable learning signals.

To address this issue, we introduce the Intermediate Memory Recall (IMR) task. After generating the $i$-th complete trajectory $o_i^E$, we randomly select an intermediate interaction step $k$. We then take the corresponding memory state $o^E_{i,k}$ together with a randomly selected previous chunk input $c_{l}$ ($l<k$).
The model is then required to recover the entity $\tilde y_i$ from the masked chunk $\tilde c_l$ within a single interaction step, following $o^I_{i,j}\sim\pi_\theta(\cdot|q^I,\tilde x_i)$, where $\tilde x_i=o^E_{i,k}\oplus\tilde c_l$ and $q^I$ is the IMR task prompt detailed in Appendix~\ref{sec:template}.

This objective explicitly encourages the model to preserve sufficient historical information within the current memory state. As a result, the learned memory representations become both information-rich and directly retrievable for downstream reasoning.

\subsection{Joint GRPO Optimization}\label{sec:optim}

We employ GRPO as the reinforcement learning algorithm.
Figure~\ref{fig:grpo} provides an overview.
For each training sample $(p_{1:N}, y)$, we first sample $G_1$ end-to-end trajectories $\{o_i^E\}_{i=1}^{G_1}$ under the current policy. Then, for each sampled trajectory $o_i^E$, we construct one IMR prompt and further sample $G_2$ IMR trajectories $\{o_{i,j}^I\}_{j=1}^{G_2}$.
We extract the answers $\hat y^E_i$ and $\hat y^I_{i,j}$ from these trajectories and compute the exact-match reward.
For the IMR task, we have:
\begin{equation}
R_{i,j}^{I}=\mathbb I[\hat y^I_{i,j}=\tilde y_i].
\end{equation}
For the end-to-end task, the reward consists of two components: the exact-match reward for the final prediction and the associated IMR rewards:
\begin{equation}
R_i^{E}=\mathbb I[\hat y^E_i=y]+\frac{\lambda}{G_2}\sum_{j=1}^{G_2}R_{i,j}^I,
\end{equation}
where $\lambda$ is a balancing coefficient.
The intuition behind this design is twofold. First, IMR rewards directly train the model to retrieve and reason over information stored in memory. Second, augmenting end-to-end rewards with IMR outcomes encourages the model to generate memory states that remain useful for future retrieval and reasoning.

Since each end-to-end trajectory consists of multiple interaction steps, we treat each step as an independent conversation instance for advantage estimation and policy optimization. Following Dr.~GRPO~\citep{liuUnderstandingR1ZeroLikeTraining2025}, we adopt the unnormalized advantage formulation:
\begin{equation}
    \hat A_{i,j,k}=R_i-{\rm mean}\{R_i\}_{i=1}^G,
\end{equation}
where $i,j$ and $k$ denote the index for trajectory, interaction step, and token, respectively. The advantage computed from the final trajectory reward is broadcast to all interaction steps.
Finally, all end-to-end and IMR samples are jointly optimized using the GRPO objective in Eq.~\eqref{eq:loss}. For notational simplicity, we omit $q^{E/I}$ and define a unified trajectory collection $o_i=(o_{i,1}^E,\cdots,o_{i,|o_i^E|}^E,o_{i,1}^I,\cdots,o_{i,G_2}^I)$,
which combines the end-to-end trajectory with its associated IMR trajectories.

\begin{equation}\label{eq:loss}
\begin{split}
    \mathcal{J}(\theta)\!=\!\E_{(p,y)\sim\train,\{o^E_i\}_{i=1}^{G_1}\sim\pi_\theta(\cdot|c),\{o_{i,j}^I\}_{j=1}^{G_2}\sim\pi_\theta(\cdot|\tilde x_i)}
    \left[\frac{1}{\sum_{i=1}^{G_1}|o_i^E|+G_1G_2}\sum_{i=1}^{G_1+G_2}\sum_{j=1}^{|o_i|}\sum_{k=1}^{|o_{i,j}|}C_{i,j,k}\right],
\end{split}
\end{equation}
\begin{equation*}
\begin{split}
    C_{i,j,k}=\min\!\Big(r_{i,j,k}(\theta)\hat A_{i,j,k},
    {\rm clip}(r_{i,j,k}(\theta),1\!-\!\varepsilon_{\rm low},1\!+\!\varepsilon_{\rm high})\hat A_{i,j,k}\Big)-\KL(\pi_{\theta}||\pi_{\rm ref})),
\end{split}
\end{equation*}
\begin{equation*}
r_{i,j,k}(\theta) = 
\begin{cases}
    \frac{\pi_\theta(o_{i,j,k}|c_j,o_{i,j,<k})}{\pi_{\rm old}(o_{i,j,k}|c_j,o_{i,j,<k})} & i\le G_1, \\
    \frac{\pi_\theta(o_{i,j,k}|\hat x_i,o_{i,j,<k})}{\pi_{\rm old}(o_{i,j,k}|\hat x_i,o_{i,j,<k})} & i>G_1.
\end{cases}
\end{equation*}

\section{Experiments}
We evaluate the effectiveness of MemTrain by measuring the final downstream performance after post-training. We consider two representative tasks: (1) long-context multi-hop question answering (\S~\ref{sec:longQA}), which closely matches the memory training setting where the model reads chunked long documents and answers questions; and (2) multi-hop question answering with search tools (\S~\ref{sec:rag}), an \textbf{out-of-domain} retrieval-augmented setting in which the model iteratively retrieves external information and performs reasoning to produce the final answer.
For post-training, we adopt~\citep{yu2025memagent} and MEM1~\citep{zhou2025MEM1LearningSynergize}, as they are the only open-source algorithms among related works.

\subsection{Memory Training Setup}
\paragraph{Dataset.}
We use the most general Wikipedia as the unsupervised corpus for memory training.
Entities are identified using the NER system provided by the spaCy library.
For each pivot passage, we retrieve the top-29 semantically related passages from the corpus and further augment them with 120 randomly sampled passages.
This process produces 30k training documents with lengths ranging from 24k to 40k tokens.

\paragraph{Implementation.}
Our training framework is implemented based on veRL~\citep{shengHybridFlowFlexibleEfficient2025}.
We adopt GRPO~\citep{deepseek-aiDeepSeekR1IncentivizingReasoning2025} with a KL regularization coefficient of $1\times10^{-3}$, and follow DAPO~\citep{yuDAPOOpenSourceLLM2025} by filtering out samples whose rewards are entirely zero or entirely one.
Following prior context memory agent works~\citep{yu2025memagent,zhou2025mem1}, we limit the context length to 8192 tokens, including 1024 tokens for instructions, 5120 tokens for input chunks, 1024 tokens for memory, and 1024 tokens for model responses.
Consequently, each input consists of at most $40k/5k=8$ chunks.
We use a batch size of 32, generate $G_1=8$ end-to-end rollouts, and sample $G_2=8$ IMR trajectories for each rollout.
Training is conducted for 300 steps with a learning rate of $1\times10^{-6}$.
The IMR coefficient $\lambda$ is set to 0.5.
For backbone model selection, we evaluate two widely used instruction models: Qwen3-4B-Instruct-2507 and Qwen2.5-7B-Instruct.

\subsection{Long-Text Multi-Hop QA}\label{sec:longQA}

\paragraph{Post-Training.}
We adopt MemAgent~\citep{yu2025memagent} as the downstream post-training algorithm.
All hyperparameters follow the settings described in the MemAgent paper.
We train for 500 steps for convergence using a rollout batch size of 32, an update batch size of 8, and a learning rate of $1\times10^{-6}$.
For each backbone, we train two variants: one directly post-trained with MemAgent and another initialized from the MemTrain checkpoint before post-training, with three different seeds.

\paragraph{Evaluation.}
We evaluate on the long-context HotpotQA benchmark introduced by~\citet{yu2025memagent}, which is specifically designed to study performance under varying context lengths.
The input length ranges from 7k to 896k tokens.
For direct evaluation of the original backbone models, the entire document is provided in a single context window.
For models trained after MemTrain or MemAgent, we adopt the chunked memory pipeline.

\paragraph{Results.}
Table~\ref{tab:hotpotqa} demonstrates that our memory training framework consistently provides substantial gains for subsequent memory-oriented post-training. Compared with directly applying MemAgent, the combination of MemTrain and MemAgent achieves significantly higher average performance on both backbone models, improving 5.17\% on Qwen3-4B-Instruct and 17.67\% on Qwen2.5-7B-Instruct. More importantly, these improvements are highly consistent across all context lengths, ranging from 7k to 896k tokens, indicating that the proposed memory training stage provides a strong initialization for downstream long-horizon memory learning.

Another notable observation is the strong length generalization ability introduced by MemTrain. Although the training context length (32k$\sim$40k) is closest to 28k, the gains transfer effectively to both substantially shorter and longer contexts. This effect is particularly evident on Qwen2.5-7B-Instruct. While MemAgent drops from 62.50\% at 28k to 41.41\% at 896k, corresponding to a decrease of 21.09\% points, MemTrain+MemAgent only decreases from 77.34\% to 68.75\%, a much smaller drop of 8.59\% points despite the 32$\times$ increase in context length.
The improvements also extend to shorter contexts such as 7k and 14k, indicating that MemTrain learns more transferable and length-generalizable memory maintenance and retrieval behaviors rather than overfitting to a specific training horizon.
Similar trends are consistently observed on Qwen3-4B-Instruct.

Furthermore, MemTrain alone already endows the model with considerable multi-turn question answering and memory capabilities, despite being trained entirely without labeled supervision. Compared with the original models, MemTrain improves the average performance from 21.97\% to 56.15\% on Qwen3-4B-Instruct and from 20.80\% to 45.41\% on Qwen2.5-7B-Instruct.

\begin{table*}[t]
\centering
\resizebox{\textwidth}{!}{
\begin{tabular}{lccccccccc}
\toprule
\multirow{2}{*}{\textbf{Model}} & \multicolumn{9}{c}{\textbf{Length}} \\ 
\cmidrule(lr){2-10}
 & \textbf{7k} & \textbf{14k} & \textbf{28k} & \textbf{56k} & \textbf{112k} & \textbf{224k} & \textbf{448k} & \textbf{896k} & \textbf{Avg} \\ 
\midrule
Qwen3-4B-Instruct & 57.81 & 51.56 & 34.38 & 10.94 & 8.59 & 4.69 & 3.91 & 3.91 & 21.97 \\ 
+MemTrain & 63.28 & 60.16 & 60.16 & 57.03 & 60.94 & 58.59 & 48.44 & 40.62 & 56.15 \\ 
+MemAgent & 70.31 & 64.06 & 71.88 & 62.50 & 64.84 & 66.41 & \textbf{64.06} & 57.03 & 65.14 \\ 
\rowcolor{gray!10}+MemTrain+MemAgent & \textbf{79.69} & \textbf{73.44} & \textbf{75.78} & \textbf{73.44} & \textbf{68.75} & \textbf{67.19} & 61.72 & \textbf{62.50} & \textbf{70.31} \\ 
\midrule
Qwen2.5-7B-Instruct & 53.12 & 51.56 & 35.16 & 13.28 & 10.16 & 1.56  & 1.56  & 0.00  & 20.80 \\
+MemTrain           & 59.38 & 55.47 & 48.44 & 46.09 & 42.19 & 38.28 & 39.84 & 33.59 & 45.41 \\
+MemAgent           & 64.06 & 67.19 & 62.50 & 59.38 & 55.47 & 50.00 & 46.88 & 41.41 & 55.86 \\
\rowcolor{gray!10}+MemTrain+MemAgent  & \textbf{76.56} & \textbf{79.69} & \textbf{77.34} & \textbf{75.00} & \textbf{70.31} & \textbf{75.78} & \textbf{64.84} & \textbf{68.75} & \textbf{73.53} \\
\bottomrule
\end{tabular}
}
\caption{Model performance for long-text QA across different context lengths.}
\label{tab:hotpotqa}
\end{table*}

\subsection{Multi-Hop QA With Search Tool}\label{sec:rag}
\paragraph{Post-Training.}
We adopt MEM1~\citep{zhou2025mem1} as the downstream post-training algorithm.
Following the original MEM1 setup, training is performed on 2-objective HotpotQA and Natural Questions, with at most 6 search turns and a length limit of 1k tokens for both model responses and retrieved search results.
We employ the same retriever and local database as MEM1, and train 200 steps until convergence using a rollout batch size of 32, an update batch size of 8, and a learning rate of $5\times10^{-7}$.
As in the long-context QA setting, we train both a directly post-trained model and a model initialized from MemTrain.

\paragraph{Evaluation.}
We evaluate on 7 challenging multi-hop QA benchmarks, including 2WikiMultiHopQA~\citep{hoConstructingMultihopQA2020a}, Bamboogle~\citep{mallenWhenNotTrust2023}, HotpotQA~\citep{yangHotpotQADatasetDiverse2018a}, TriviaQA~\citep{joshiTriviaQALargeScale2017}, Natural Questions, PopQA, and MusiQUE~\citep{trivediMuSiQueMultihopQuestions2022a}.
Following the MEM1 implementation, we augment the evaluation set into a two-objective setting and report exact-match accuracy averaged across the two objectives.

\begin{table*}[t]
\centering
\resizebox{\textwidth}{!}{%
\begin{tabular}{lcccccccc}
\toprule
\textbf{Model} & \textbf{TrivalQA} & \textbf{Bamboogle} & \textbf{HotpoQA} & \textbf{NQ} & \textbf{PopQA} & \textbf{2WiKi} & \textbf{MusiQUE} & \textbf{Avg} \\
\midrule
Qwen3-4B-Instruct-2507 & 42.71 & 21.78 & 18.94 & 19.92 & 21.81 & 14.36 & 4.76 & 20.61 \\
+MEM1                  & 44.29 & 23.39 & 18.80 & 21.97 & 23.62 & 12.80 & 5.63 & 21.50 \\
\rowcolor{gray!10}+MemTrain+MEM1         & \textbf{55.63} & \textbf{34.68} & \textbf{27.85} & \textbf{32.24} & \textbf{37.91} & \textbf{25.84} & \textbf{10.43} & \textbf{32.08} \\
\midrule
Qwen2.5-7B-Instruct & 18.84 &  8.87 & 11.15 & 12.22 & 12.59 & 10.45 &  4.43 & 11.22 \\
+MEM1                  & 49.08 & 22.58 & 19.79 & 24.21 & 27.13 & 17.81 &  6.96 & 23.94 \\
\rowcolor{gray!10}+MemTrain+MEM1         & \textbf{57.21} & \textbf{30.65} & \textbf{27.73} & \textbf{35.18} & \textbf{38.36} & \textbf{27.32} & \textbf{10.64} & \textbf{32.44} \\
\bottomrule
\end{tabular}%
}
\caption{Model performance for multi-hop QA with search tools across different benchmarks.}
\label{tab:rag}
\end{table*}

\begin{figure*}[t]
    \centering
    \includegraphics[width=0.95\linewidth]{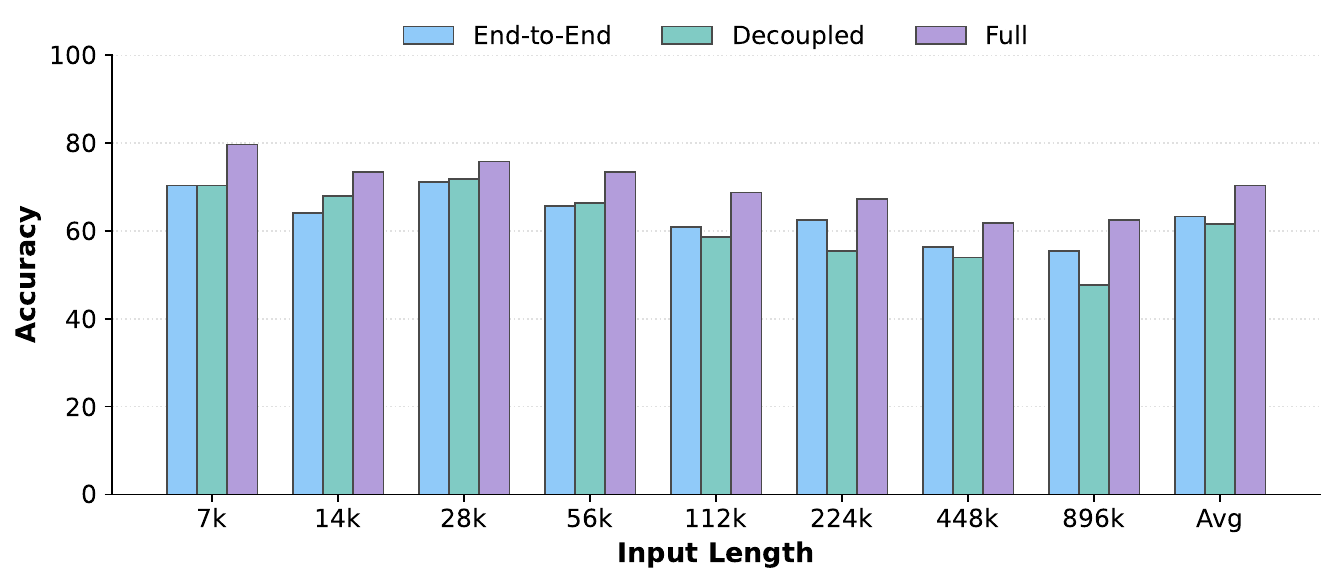}
    \caption{Ablations results on long-context HotpotQA across different context length.}
    \label{fig:ablation}
\end{figure*}

\paragraph{Results.}
Table~\ref{tab:rag} shows that MemTrain generalizes well to search-based multi-hop QA despite a clear distribution shift from memory training.
Across models, MemTrain+MEM1 consistently improves over MEM1 on all benchmarks.
On Qwen3-4B-Instruct-2507, the average performance increases by 10.58 points, and on Qwen2.5-7B-Instruct by 8.50 points.
MemTrain-only models are not involved in comparison because the they are not exposed to tool-use environment.

The improvements are consistent across datasets and are more pronounced on harder multi-hop tasks. In particular, the largest gains are observed on PopQA, NQ, and 2Wiki, with improvements of +11.23, +10.97, and +9.51 on Qwen2.5-7B-Instruct, and +14.29, +10.27, and +13.04 on Qwen3-4B-Instruct-2507, respectively. This may be attributed to the fact that these tasks require maintaining and integrating a larger number of intermediate evidences across retrieval steps, where improved memory construction and utilization from memory training becomes more critical.
Notably, on MusiQUE, directly applying MEM1 yields only marginal improvements over the base model (e.g., +2.53 on Qwen2.5-7B-Instruct), whereas incorporating MemTrain leads to a much larger gain (+6.21), suggesting that memory-aware training is particularly beneficial in more retrieval-sensitive settings.

\section{Analysis}

\begin{table*}[t]
\centering
\small
\begin{tabular}{p{0.47\textwidth}|p{0.47\textwidth}}
\toprule
\textbf{MemAgent} & \textbf{MemTrain+MemAgent} \\
\midrule
\multicolumn{2}{c}{\textbf{Question}} \\
\multicolumn{2}{p{0.94\textwidth}}{\textit{The director of the romantic comedy ``Big Stone Gap'' is based in what New York city?}} \\
\midrule
\multicolumn{2}{c}{\textbf{Chunk 1}} \\
\multicolumn{2}{p{0.94\textwidth}}{(Documents 1--36 omitted) \ldots Document 37: Adriana Trigiani is an Italian American best-selling author, television writer, \underline{film director}, and entrepreneur \underline{based in Greenwich Village, New York City}.} \\
\midrule
\textbf{Memory 1} & \textbf{Memory 1} \\
\ldots{}Document 37 mentions Adriana Trigiani, an Italian American best-selling author, television writer, \textcolor{red}{\textbf{film director}}, and entrepreneur \textcolor{red}{\textbf{based in Greenwich Village, New York City}}. While she is a film director and based in New York City, there is no direct evidence that she directed ``Big Stone Gap''\ldots & \ldots{}The problem asks about the director of ``Big Stone Gap.'' However, after reviewing the provided section, \textcolor{red}{\textbf{no film titled ``Big Stone Gap'' is mentioned}}. The only romantic comedy films discussed are ``Nola'' and ``Just Another Romantic Wrestling Comedy''\ldots \\
\midrule
\multicolumn{2}{c}{\textbf{Chunk 2}} \\
\multicolumn{2}{p{0.94\textwidth}}{(Documents 38--43 omitted) \ldots Document 44: \textit{Big Stone Gap} is a 2014 American drama romantic comedy film \underline{written and directed by Adriana Trigiani} and produced by Donna Gigliotti\ldots (Documents 45--50 omitted)} \\
\midrule
\textbf{Memory 2} & \textbf{Memory 2} \\
\ldots{}Document 44 confirms ``Big Stone Gap'' (2014) was \textbf{written and directed by Adriana Trigiani}. Document 37 confirms Adriana Trigiani is \textbf{based in Greenwich Village, New York City}. Thus, the director is based in \ldots & \ldots{}Document 44 confirms ``Big Stone Gap'' (2014) was written and directed by Adriana Trigiani. \textcolor{red}{\textbf{However, no information about her base location is available in memory.}} Thus, the required information is not present\ldots \\
\midrule
\multicolumn{2}{c}{\ldots}\\
\midrule
\textbf{Final Answer} & \textbf{Final Answer} \\
\textcolor{blue}{\textbf{Greenwich Village}} & \textcolor{red}{\textbf{Cannot be determined}} \\
\bottomrule
\end{tabular}
\caption{Case study comparing models with and without MemTrain. Key information is underlined in the input chunks and highlighted in bold within the memory. Critical differences are marked in \textcolor{red}{red}.}
\label{tab:case_study}
\end{table*}

\begin{figure}[t]
    \centering
    \includegraphics[width=0.80\linewidth]{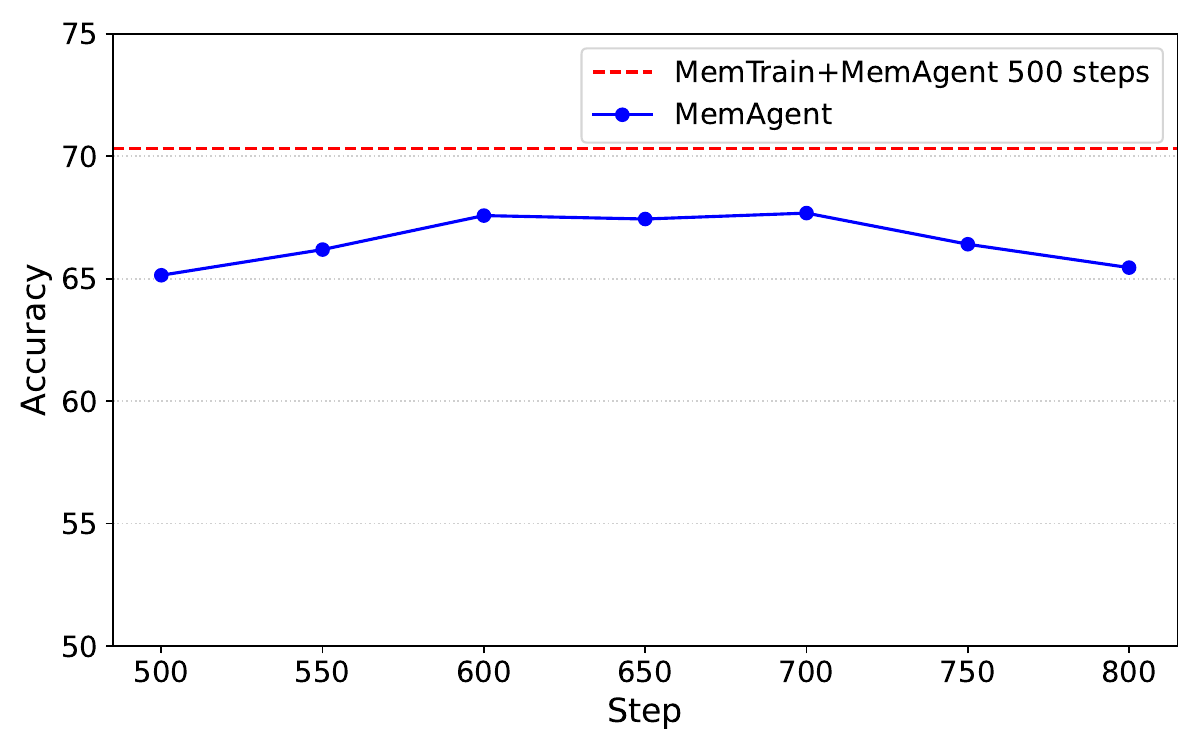}
    \caption{Performance comparison between MemTrain and continual post-training.}
    \label{fig:scaling}
\end{figure}

\subsection{Ablation Study}
To further investigate the contribution of each component in MemTrain, we design two ablation variants:
(1) \textbf{End-to-End}, which removes the IMR branch and retains only the end-to-end prediction objective; and
(2) \textbf{Decoupled}, which computes rewards for end-to-end trajectories solely based on final prediction correctness, decoupled from IMR.

As shown in Figure~\ref{fig:ablation}, the Full model consistently outperforms both ablation variants across all evaluated context lengths, demonstrating the importance of IMR.
Specifically, removing the IMR branch decreases the average score from 70.31\% to 63.28\%.
This degradation consistently appears across all context lengths, indicating that the end-to-end prediction objective alone does not provide sufficient supervision for identifying and preserving critical information throughout extremely long interaction histories.

Compared with the End-to-End variant, the Decoupled variant achieves stronger performance on relatively shorter contexts ($\leq56k$), suggesting that IMR learning improves memory utilization.
However, its performance deteriorates significantly as the context length increases.
One possible explanation is that the decoupled objective fails to provide sufficient guidance for high-quality memory generation, forcing the model to solve tasks based on poorly constructed memories and consequently leading to more severe hallucination under long-horizon settings.

\subsection{Memory Training V.S. Post-Training Scaling}
In this section, we compare the gains brought by memory training with those obtained from simply scaling post-training.
Starting from the MemAgent checkpoint at step 500 on Qwen3-4B-Instruct-2507, we continue post-training for an additional 300 steps.
We report the average accuracy across all input lengths.

As shown in Figure~\ref{fig:scaling}, post-training is already close to saturation after step 500, and further scaling yields only marginal improvements or even performance degradation.
Even at the best-performing checkpoint around step 700, the model initialized with MemTrain still maintains an advantage of 2.64 percentage points.
These results suggest that although memory training introduces additional computational cost, it effectively raises the performance ceiling of downstream post-training in a manner that cannot be replicated by simply extending post-training.
Therefore, allocating additional GPU resources to memory-oriented training appears to be a meaningful investment.

\subsection{Case Study}
We present a representative case of Qwen3-4B-Instruct-2507 to understand the effect of MemTrain.
As shown in Table~\ref{tab:case_study}, direct MemAgent fails to retain the critical information at the memory update step after chunk 1, resulting in an inability to answer despite finding the director's identity in chunk 2.
MemTrain successfully preserves the key entity information (Adriana Trigiani's location) in memory from chunk 1, enabling correct answer deduction in chunk 2.

\section{Conclusion}
In this work, we introduce MemTrain, the first self-supervised memory training framework for improving the general-purpose memory capability of LLMs.
We design two coupled proxy tasks—end-to-end masked reconstruction and intermediate memory recall—to jointly encourage memory completeness, faithful compression, and effective utilization. 
We perform memory training on Wikipedia corpora and demonstrate consistent improvements on downstream long-text and search-based question answering tasks across two models.

\bibliography{colm2026_conference}
\bibliographystyle{colm2026_conference}

\newpage
\appendix
\onecolumn
\begin{figure*}[t]
\begin{promptbox}{\textbf{End-to-End Memory Generation Prompt}}
You are presented with a problem, a section of an article that may contain the answer to the problem, and a previous memory. Please read the provided section carefully and update the memory with the new information that helps to answer the problem. Be sure to retain all relevant details from the previous memory while adding any new, useful information.
<problem>  Based on the entire article, which entity is represented by the {MASK} placeholder? </problem>
<memory> {memory} </memory>
<section> {chunk} </section>

Updated memory:
\end{promptbox}
\begin{promptbox}{\textbf{End-to-End Answer Generation Prompt}}
You are presented with a problem and a previous memory. Please answer the problem based on the previous memory and put the answer in boxed{}.
<problem>  Based on the entire article, which entity is represented by the {MASK} placeholder? </problem>
<memory> {memory} </memory>
<section> {section} </section>
\end{promptbox}
\begin{promptbox}{\textbf{Intermediate Memory Recall Prompt}}
You are presented with a memory from previously read sections, and a section containing placeholder [TARGET]. Based on the memory, please predict the entity marked by [TARGET] and put the answer in boxed{}.
<memory> {memory} </memory>
<section> {masked chunk} </section>
\end{promptbox}
\end{figure*}

\section{Prompt Template}
\label{sec:template}

MemTrain employs three prompt templates, as illustrated below.
For the end-to-end masked reconstruction task, we adopt the prompt design from MemAgent~\citep{yu2025memagent} and set the problem as a fixed masked prediction instruction.
Specifically, the memory generation prompt is applied iteratively until all text chunks have been processed, after which the answer generation prompt is used to produce the final output.
For the intermediate memory recall task, we introduce the placeholder [TARGET] to distinguish it from [MASK], thereby preventing the LLM from being confused about which reconstruction objective to perform.

\end{document}